\documentclass[11pt, letterpaper]{arxiv}
\usepackage[all]{hypcap}
\usepackage[comma,numbers,sort,compress]{natbib}
\usepackage{hyperref}[citecolor=magenta]
\usepackage[capitalize,noabbrev]{cleveref}
\hypersetup{
    colorlinks = true,
    citecolor = {magenta},
}

\usepackage{microtype}
\usepackage{graphicx}
\usepackage{subfigure}
\usepackage{booktabs} 

\usepackage{xcolor}
\usepackage{colortbl}
\usepackage{algorithm}
\usepackage{algpseudocode}
\usepackage{setspace}

\usepackage{amsmath}
\usepackage{amssymb}
\usepackage{mathtools}
\usepackage{amsthm}

\usepackage{placeins}

\usepackage{fleqn, tabularx}

\usepackage{float}
\usepackage[most,skins,theorems]{tcolorbox}
\tcbset{
  aibox/.style={
    width=\linewidth,
    top=7pt,
    bottom=2pt,
    colback=rliableolive!13!white,
    colframe=black,
    colbacktitle=black,
    enhanced,
    center,
    attach boxed title to top left={yshift=-0.1in,xshift=0.15in},
    boxed title style={boxrule=0pt,colframe=white,},
  }
}
\newtcolorbox{AIbox}[2][]{aibox,title=#2,#1}

\usepackage{wrapfig}
\definecolor{lightblue}{rgb}{0.22,0.45,0.70}
\usepackage{fleqn, tabularx}

\definecolor{rliableolive}{HTML}{BBCC33}
\definecolor{rliableblue}{HTML}{77AADD}
\definecolor{rliablered}{HTML}{EE8866}

\usepackage{crossreftools}
\usepackage[suppress]{color-edits}

\usepackage{dsfont}
\usepackage{nicefrac}
\newtcolorbox{analysisbox}[1][]{
    enhanced jigsaw,
    colback=white,
    colframe=blue!75!black,
    fonttitle=\bfseries,
    boxsep=5pt,
    left=5pt,
    right=5pt,
    top=5pt,
    bottom=5pt,
    title=#1,
}
\definecolor{editInitialResponse}{RGB}{255, 235, 156} 
\definecolor{editBacktrack}{RGB}{0, 0, 139}
\definecolor{editRevisedResponse}{RGB}{255, 182, 193}

\usepackage{pifont}
\usepackage{soul}
\definecolor{highlightmistake}{RGB}{255, 179, 179} 
\definecolor{highlightcorrect}{RGB}{179, 255, 179}

\usepackage[capitalize,noabbrev]{cleveref}

\theoremstyle{plain}

\theoremstyle{definition}

\theoremstyle{remark}

\usepackage[textsize=tiny]{todonotes}
\usepackage{multirow}
\usepackage{enumitem}

\usepackage{amsmath,amsfonts,bm}

\def\eqref#1{Eq.~\ref{#1}}

\def\1{\bm{1}}

\DeclareMathAlphabet{\mathsfit}{\encodingdefault}{\sfdefault}{m}{sl}
\SetMathAlphabet{\mathsfit}{bold}{\encodingdefault}{\sfdefault}{bx}{n}

\newcommand{\E}{\mathbb{E}}

\newcommand{\bz}{\mathbf{z}}

\newcommand{\by}{\mathbf{y}}

\newcommand{\bx}{\mathbf{x}}

\newtcolorbox{promptbox}[2][]{  
listing only,
enhanced,
breakable,
colback=rliableolive!13!white,
colframe=black,
fontupper=\ttfamily,
title=#2,
#1}

\newcommand{\method}{{\texttt{CaRT}}}

\title{\method: Teaching LLM Agents to Know When They Know Enough}

\author[*1]{Grace Liu}
\author[*1]{Yuxiao Qu}
\author[1]{Jeff Schneider}
\author[1]{Aarti Singh}
\author[1]{Aviral Kumar}

\affil[1]{Carnegie Mellon University}
\affil[*]{Equal contribution}

\correspondingauthor{gliu2@andrew,cmu.edu, yuxiaoq@andrew.cmu.edu \newline \textbf{Project Website:} \textnormal{\url{https://graliuce.github.io/cart-page/}}}

\begin{document}

\maketitle

\begin{figure}[h]
\vspace{-0.8cm}
\centering
\includegraphics[width=0.99\linewidth]{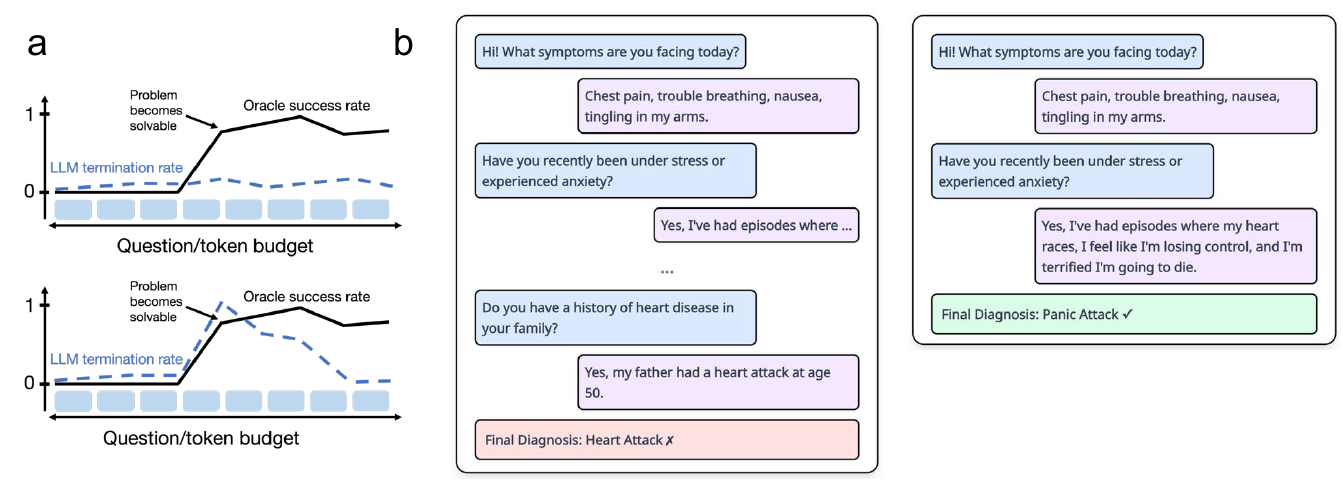}
\vspace{-0.2cm}
\caption{\footnotesize{\textbf{\emph{A schematic illustration of the termination behavior of models}} with and without our proposed approach. While LLMs typically fail to recognize the best points to stop thinking or questioning often either overshooting or undershooting the amount of information needed (a - top, b - left), our approach \method{} imbues them with the ability to correctly identify this point.}}
\vspace{-0.2cm}
\label{fig:cart_teaser}
\end{figure} 

{\absfont \textbf{Abstract:} Many tasks require learned models to strategically gather relevant information over multiple rounds of interaction before actually acting on a task. Strategic information gathering requires models to know not only how to effectively acquire information, but also when to stop gathering information and make a decision, in order to avoid overthinking or getting derailed when acting. In this paper, we formalize this problem and introduce Counterfactuals and Reasoning for Termination (\method{}), an approach for teaching LLMs when to stop seeking information.  To appropriately learn when to terminate, \method{} fine-tunes LLMs using counterfactual pairs of trajectories, one where termination is appropriate and a minimally modified version of the same trajectory where it is not. It trains the LLM to explain the rationale for the termination decision in either case via verbal reasoning, and imbues this capability into the base LLM via fine-tuning. We instantiate \method{} in two domains: interactive medical diagnosis and math problem solving. In both domains, we find that \method{} improves the efficiency of information gathering and task success rate compared to other fine-tuning methods.}

\vspace{-0.3cm}
\section{Introduction}
\label{intro}
\vspace{-0.2cm}

Strategic information gathering is core to problem solving and decision making with AI~\citep{thrun1992efficient}. 
For example, when attempting to design and prescribe a course of treatment to a user, it is important to gather complete information about their symptoms.
In many scenarios, information gathering relies not only on deciding how to acquire more information, but also on deciding \emph{when the model has gathered enough information} to solve the task. A model that stops too late wastes resources, while one that stops too early risks failure. Moreover, additional information can be detrimental for many of the transformer architecture based models used commonly right now, where extra information in the input context may lead the model to latch onto spurious information~\citep{liu2023lost,wang2023primacy}. The ability to recognize when ``I know enough'' is therefore essential to efficient and reliable problem solving.

Deciding when to stop thinking, interacting, or seeking information is challenging because it requires predicting the expected future utility of continuing under the model's current policy. Classical statistical approaches to this problem rely on accurately estimating a value function~\citep{nie2019learning, thomas2016data} with limited data and typically operate in domains with well-defined environment dynamics, such as airline ticket purchasing~\citep{groves2015optimizing,goel2017sample}. 
We study this problem of termination when large language models (LLMs) are utilized as decision makers: LLMs possess natural language capabilities and rich priors about the world, providing the potential for greater versatility, domain applicability, and generalization compared to classical statistical approaches on this problem. However, off-the-shelf LLMs even struggle to accurately predict their probability of success~\citep{savage2024large, omar2025benchmarking, sun2025large, groot2024overconfidence} and are unable to conduct principled exploration~\citep{arumugam2025toward}. These limitations put into question whether current recipes for training LLMs imbue them with the ability to quantify the value of what they don't know, a key skill for effective termination.

In this paper, we build an approach to imbue LLMs with the ability to stop or ``terminate'' their internal thinking processes and/or environment interaction at the right point for maximal performance, without wasting computation or interaction. Our key insight is that textual reasoning itself can be used to learn accurate and generalizable termination behavior, as long as this reasoning is done comparatively (and contrasts the benefits of terminating and continuing). Our approach, \textbf{Counterfactuals and Reasoning for Termination} (\method), fine-tunes models with \emph{counterfactual} pairs: trajectories where termination is appropriate and minimally modified trajectories where it is not, combined with explicit natural language reasoning traces that justify why termination is the right decision. This comparative reasoning signal enables models to implicitly implement a ``verbalized'' value function, allowing the LLM to foresee the benefits of termination or continuation, compare them, and choose the better of the two decisions. We instantiate \method{} in two multi-step domains: \textbf{a)} interactive medical diagnosis, which requires interaction with an external environment, and \textbf{b)} mathematical reasoning, which requires spending test-time compute to think longer for harder problems. Across both of these domains, we find that \method{} demonstrates superior termination behavior compared to the base model and SFT approaches.

Our contributions are:
\textbf{(1)} We develop an approach for studying and formalizing the problem of optimal information seeking in long chain-of-thought and multi-turn settings. \textbf{(2)} We demonstrate that training with CaRT improves termination behavior for both medical diagnosis and math problem solving tasks. \textbf{(3)} We show that off-the-shelf LLMs fail to terminate efficiently. We analyze the advantages provided by training with reasoning and counterfactuals through ablations and representation analysis.
\vspace{-0.3cm}
\section{Related Work}
\vspace{-0.2cm}

\textbf{Learning when to act.} The challenge of deciding when to act versus when to continue gathering information is central to dynamic decision-making problems~\citep{moodie2007demystifying, nie2019learning}. In this setting, a decision maker can provide a solution only once but can decide when to provide that solution as more information becomes available. This problem setting appears across diverse domains, including timing of medical treatment~\citep{when2009timing,prasad2017reinforcement,gottesman2019guidelines, ajdari2019towards}, social interventions~\citep{durlak2011impact}, and natural resource harvesting~\citep{behringer2014optimal, pascoe2002optimal}. 

Statistical approaches to optimal termination have focused on learning stopping rules and developing sample-efficient estimators~\citep{nie2019learning, liu2018augmented,goel2017sample,thomas2016data}, while empirical methods have applied deep learning to improve value estimation in high-dimensional environments~\citep{becker2019deep,felizardo2022solving}.
However, these methods typically operate in settings with well-defined environmental dynamics or features crafted with domain knowledge~\citep{huang2022innermonologueembodiedreasoning}, limiting their open-ended applicability. For more open-ended problem settings, LLMs provide the versatility of operating with natural language. Additionally, LLMs possess rich priors about the world and flexible thinking capabilities: they can simulate possible futures through chain-of-thought~\citep{yan2024efficientreinforcementlearninglarge} and adapt policies to new tasks without explicit environment models~\citep{babu2025adaptivedomainmodelinglanguage}
Despite these advantages, LLMs face limitations for learning effective termination. Prior work shows that optimal termination depends on accurate value estimation~\citep{nie2019learning,liu2018augmented,goel2017sample,thomas2016data}. However, off-the-shelf LLMs struggle to accurately predict their probability of success, even in fairly simple, single-turn settings~\citep{savage2024large, omar2025benchmarking, sun2025large, groot2024overconfidence} 
and exhibit inefficient exploration in sequential bandit environments~\citep{nie2024evolve,krishnamurthy2024can,arumugam2025toward}. 
Our approach builds on the versatility of LLMs but addresses their shortcomings at estimating the future by training LLMs with counterfactual examples and explicit reasoning for termination in multi-step tasks. 

\textbf{Information seeking with LLMs.} When answering user queries, standard LLMs typically provide an answer without seeking additional information. Even systems such as OpenAI Deep-Research ask only one clarifying question. These fairly static approaches are suboptimal because LLMs often produce answers even when a query is underspecified or missing critical details~\citep{feng2024don, zhang2024clamber}. To improve information seeking, prior work has developed methods for LLMs to detect ambiguity and ask clarifying questions before answering~\citep{deng2023plug, deng2023prompting, zhang2023clarify,li2023eliciting,pang2024empowering}. More recent work has extended this to multi-turn settings, such as medical diagnosis~\citep{jia2025ddo}, where agents may gather several pieces of information before providing a recommendation. These systems often improve the quality of questions through SFT or reinforcement learning~\citep{li2025aligning, zhu2025ask, zhang2024ask, chopra2025feedback}. Some systems maintain a separate confidence module to inform the decision maker~\citep{jia2025ddo, bani2025language} but do not explicitly optimize for termination, which our approach aims to do directly.




\textbf{Teaching LLMs to terminate optimally.} Beyond deciding \emph{which} question to ask, an information seeker must decide \emph{when} to stop asking questions and terminate. Methods for addressing termination in single-shot or few-shot settings include prompting or using rollout diversity to measure user ambiguity~\citep{zhang2023clarify, pang2024empowering, deng2023prompting, kuhn2022clam} and preference fine-tuning~\citep{zhang2024modeling, wang2024learning}. Preference fine-tuning approaches can improve termination capability in single-step settings, but these approaches are limited because question ambiguity is often subjective and high-quality human annotations are expensive to collect. Recent work in the domain of math reasoning with LLMs explores training LLMs with length penalties or constraining them to reason with only short traces~\citep{yeo2025demystifyinglongchainofthoughtreasoning,muennighoff2025s1simpletesttimescaling,arora2025traininglanguagemodelsreason}. While such approaches can reduce over-reliance on unnecessarily long reasoning, they suffer from poor adaptivity: models trained under strict length constraints struggle to generalize to out-of-distribution tasks, limiting flexibility across diverse problem structures is essential.

Our approach differs from prior work by directly optimizing for termination behavior in long-form, multi-step reasoning tasks rather than one-shot or few-shot ambiguity detection~\citep{zhang2023clarify,zhang2024modeling}. Instead of relying on subjective labels for when clarification is needed~\citep{wang2024learning}, we use the model's downstream task success rate at each timestep as a dense, scalable reward signal to learn to decide when to terminate, by learning a sort of an ``implicit value function'' that balances exploration and termination. 

\vspace{-0.3cm}
\section{Problem Statement and Notation}
\vspace{-0.25cm}

The process of attempting to gain information, even if those actions may be unrewarding in the short-term, is typically referred to as \emph{exploration}~\citep{ladosz2022exploration}. We use the terms \emph{explicit} vs. \emph{implicit} information seeking to refer to whether information gained during exploration comes from an external agent (e.g. a human providing an answer to a question) or whether the information gained comes from the model's internal dialogue (e.g. internal thinking of the model generated by spending more test-time compute~\citep{setlur2025e3}). For implicit information seeking, we segment seeking behavior into steps of reasoning, where the end of a step of reasoning provides a checkpoint for termination.
  
We are given a dataset $\mathcal{D}_\text{train} = \{(x_i, y_i^*)\}_{i=1}^N$ of problems $x_i$ and corresponding oracle answers $y_i^*$. For explicit information seeking, we also assume access to an environment $\mathcal{E}$ that, given a query, 
returns an observation $\mathbf{o}$ (e.g., feedback, retrieved information) from the environment, and a reward function $r(x, y)$ that measures the quality of a final answer $y$ relative to the ground truth.

\begin{wrapfigure}{r}{0.49\textwidth}
\vspace{-0.1cm}
\centering
\includegraphics[width=0.95\linewidth]{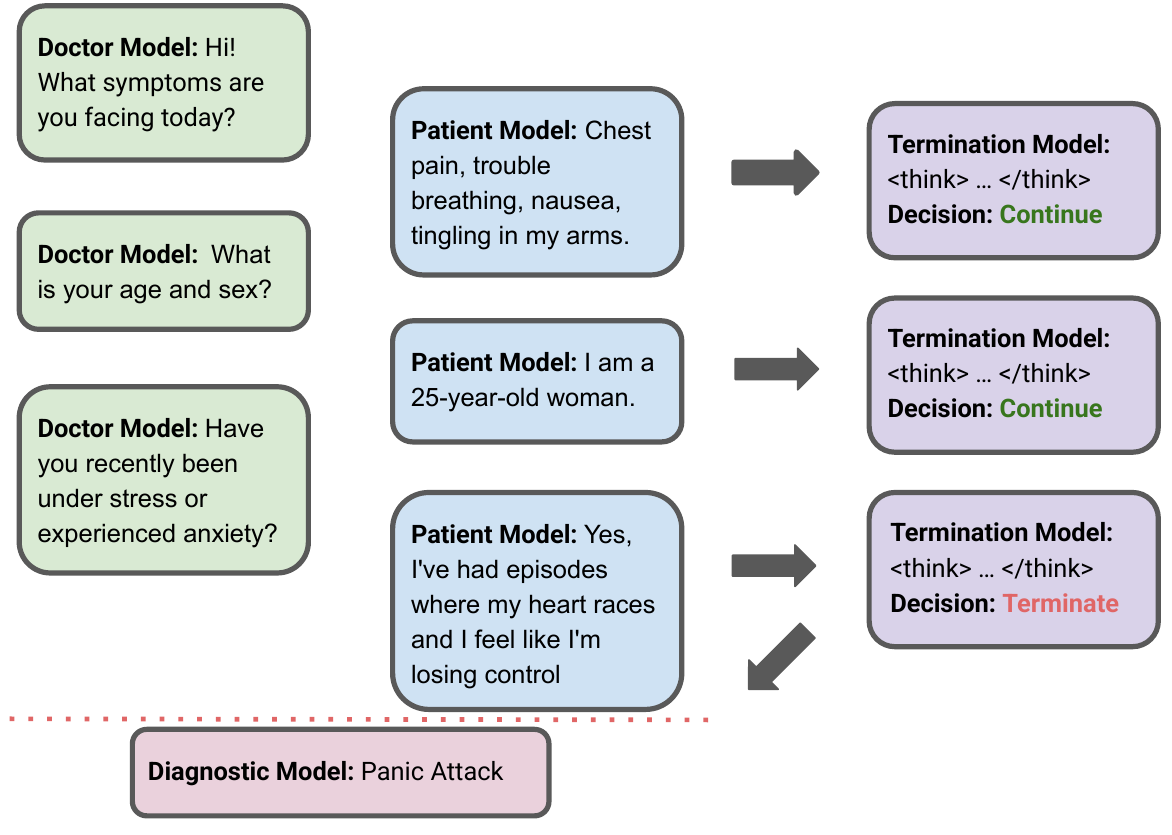}
\vspace{-0.2cm}
\caption{\footnotesize{\emph{\textbf{{An example of terminating information gathering in the medical diagnosis domain.}}}The model should terminate when there is sufficient information.}}
\label{fig:med_example}
\vspace{-0.5cm}
\end{wrapfigure} 
The LLM acts as a policy $\pi$ choosing an action $a_t \in \{\texttt{continue}, \texttt{terminate}\}$ that determines whether the model continues to seek information (explore) or reports its final answer (exploit). Each time the model chooses the continue action, it receives a stream of intermediate thinking tokens in the implicit setting\footnote{Note that these tokens correspond to the model's own context so far, though formalizing this process as ``receiving'' a stream of tokens allows us to unify terminology in all settings. Moreover, as \citet{setlur2025opt} argues, even spending more test-time compute provides information gain, making this formulation natural even in the absence of external interaction.} or environmental feedback in the explicit setting $ \bz = (o_0, a_0, o_1, ...)$. At each step, the model chooses whether to continue seeking information or to terminate and produce a final answer $\by$. The state includes prompt $\bx$, the tokens so far $\bz_{0:t}$, and observations $\mathbf{o}_{0:t}$. 
The goal is to train a policy $\pi(a_t \mid x, \bz_{0:t}, \mathbf{o}_{0:t})$ that adaptively decides when to terminate its reasoning or interaction process once sufficient information has been gathered to solve a task, balancing task accuracy with computation or interaction cost.

\begin{AIbox}{Problem Statement}
\label{def:termination_problem}
Given a problem $x \sim \mathcal{D}$, an information-seeking process producing a sequence of observations $\mathbf{o}_{0:t}$ and intermediate reasoning tokens $\bz_{0:t}$, and a policy $\pi(a_t \mid x, \bz_{0:t}, \mathbf{o}_{0:t})$ that chooses whether to continue or terminate at each step, we define the objective of adaptive termination as:
{
\setlength{\abovedisplayskip}{3pt}
\setlength{\belowdisplayskip}{3pt}
\begin{align}
\label{eq:task_objective}
\max_{\pi} \;
\mathbb{E}_{x \sim \mathcal{D}} \Bigg[
    \sum_{t=1}^T
    \mathbb{E}_{(\mathbf{o}_t, a_t) \sim \mathcal{E} \times \pi}
    \left[
        \gamma^t \, \mathbf{1}\{a_t = \texttt{terminate}\} \cdot r(x, y_t)
    \right]
\Bigg],
\end{align}
}where $\gamma \in (0,1]$ penalizes excessive computation or queries and $T$ is the maximum number of reasoning or interaction steps.
\end{AIbox}


Figure~\ref{fig:med_example} provides a simplified example to illustrate our setup in the medical question-answering setting. The problem $x$ is a medical diagnosis task with the ground truth diagnosis of panic attack. At each timestep $t$, the termination LLM receives as input the ongoing conversation and chooses $a_t \in \{\texttt{continue}, \texttt{terminate}\}$. If the action is ``continue", then the LLM will receive an additional question-answer pair $o_t$ at the next timestep. If the action is ``terminate", an external diagnostic model provides a diagnosis $y_t$ given the conversation up to that timestep. The reward $r(x, y_t)$ is then determined given the task ground truth answer and predicted diagnosis. 
\vspace{-0.4cm}
\section{\method: Counterfactuals \& Reasoning for Termination}
\vspace{-0.2cm}

Learning to terminate at the right point requires models to reason about both external and internal factors: the model must assess whether it has sufficient information to succeed and whether continued exploration is likely to be beneficial. Therefore, the model would have to accurately assess the value of currently available information and estimate the value of missing information. In order to accurately estimate the value of future information, the model must learn what information is likely to be gained through future interactions. However, because the number of potential futures is exponential (and infinite), the key challenge lies in learning effective termination behavior from limited data. Our approach, \method{}, addresses this challenge by constructing training data to include hard negative counterfactual examples and explicit reasoning traces, explaining the termination decision. Hard negative counterfactual examples are especially informative because they isolate the specific piece of information (e.g. question-answer pair) necessary to solve the task. Training models to utilize reasoning to verbalize the utility of both available and missing information before making a termination decision serves as an implicit value function and helps internalize the decision. Our approach utilizes these two components as we discuss next.

\vspace{-0.4cm}
\subsection{Component 1: Generating Hard Negative Counterfactuals}
\vspace{-0.3cm}
A na\"ive approach to teaching a model when to terminate would be to perform supervised fine-tuning (SFT) on termination labels extracted from successful information-seeking trajectories. However, this strategy is fundamentally flawed: termination decisions in real interactions are often confounded by spurious correlations. For example, a model might learn to terminate merely when the dialogue is long, or when the preceding utterance sounds confident, rather than when the necessary information has actually been gathered. \emph{How do we prevent the model from learning these shortcuts?}

Our approach draws inspiration from recent work on counterfactual data augmentation in classification~\citep{gui2025mitigating, chang2021towards, kaushik2019learning, bae2025salad, feder2023data}, which shows that providing explicitly contrasted positive and negative examples is highly effective for breaking spurious correlations. We adapt this idea to the termination problem by asking: What minimal change to the context causes a good termination decision to become a bad one (or vice versa)?
Concretely, for every trajectory where terminating is correct (i.e., leads to high answer success), we construct a hard negative counterfactual where termination would be suboptimal, ensuring that the only difference between the pair is the presence or absence of genuinely necessary information. This isolates the true causal signal the model should attend to.
Our procedure consists of:
\begin{enumerate}[leftmargin=2.0em,itemsep=4pt]
\vspace{-0.1cm}
\item \textbf{Trajectory selection}: We first identify examples of optimal termination in our dataset. Our key idea is to locate a \emph{breakpoint} in the middle of a trace (e.g., a specific question-answer pair in an interactive setting or an intermediate reasoning step in a non-interactive setting) where choosing to terminate versus continue results in a sharp change in task success. As discussed in our experiments, in the medical diagnosis setting we select prefixes where a question-answer pair yields a $\geq 50\%$ increase in success rate, while in the math reasoning setting we select prefixes where terminating yields higher success than continuing to think further.

\item \textbf{Counterfactual generation}: Our objective is to construct contrasting trajectories that contain nearly identical information but lead to opposite outcomes. To achieve this, we generate a negative (non-terminating or unsuccessful) counterpart for every termination decision by minimally altering the trace so that it no longer leads to success. Concretely, in the interactive information-seeking setting (e.g., medical diagnosis), we perturb only the final question-answer pair and find an alternate question such that the resulting success rate drops below $30\%$. This yields counterfactual pairs that differ by a single interaction yet produce maximally divergent outcomes. In the non-interactive setting (i.e., math reasoning), modifying just one single reasoning step is often not enough to reliably change success. Therefore, we find a larger sequence of reasoning steps such that removing them and terminating early reliably reduces the success rate at the task. In both cases, continuing is the best decision at the newly-generated counterfactual, while termination is the best decision originally.

\item \textbf{Contrastive labeling}: The original successful examples are labeled with a ``terminate'' decision, while the negative counterfactual examples are labeled with a ``continue'' decision. 
\end{enumerate}

\vspace{-0.4cm}
\subsection{Component 2: Verbal Reasoning for Sample-Efficient Learning From Counterfactual Data}
\vspace{-0.3cm}
We then augment the training examples found above with explicit reasoning traces that explain the termination decision. This approach is motivated by prior work showing that chain-of-thought reasoning~\citep{wei2022chain} improves generalization in a variety of settings~\citep{setlur2025opt,qu2025optimizingtesttimecomputemeta,kim2025reasoning}, but we show that it can also help improve the accuracy and generalization of implicitly learning a utility of continuing in the future.

Given the trajectory history and the termination decision for each training example, we prompt an off-the-shelf LLM (GPT-4o in our case) to generate a reasoning trace explaining why the current state warrants the termination decision. These reasoning traces serve a similar role as a value function, in that they help the model predict the best action (\texttt{terminate} / \texttt{continue}) by reasoning about potential implications of each before actually executing the action. Mechanistically, reasoning before predicting the decision makes it easier to classify the state into terminate or continue. Additionally, reasoning traces make the model's termination decisions more transparent, improving explainability which might be critical in certain domains. The combination of counterfactual data generation and reasoning generates data that teaches models to recognize and justify indicators of information sufficiency, leading to more reliable termination behavior in multi-turn information-seeking tasks.


\textbf{Training details.} We perform supervised fine-tuning (SFT) on the counterfactual examples described above. This approach performs behavioral cloning on trajectories that terminate at high-reward points and continue at low-reward points, effectively optimizing for the policy objective in Equation~\ref{eq:task_objective}. Because our counterfactual pairs isolate the most critical information determining success, we can learn this policy efficiently with limited data. For a variant of our method in the medical setting, we also perform additional RL, using GRPO~\citep{shao2024deepseekmath} on top of the fine-tuned model.  The RL training uses the same dataset with a binary reward function:  $+1$ when the model correctly terminates (success rate $\geq 0.5$) or continues (success rate $< 0.5$), and $-1$ otherwise.

\begin{AIbox}{Summary: The Main Ingredients of \method{}}
\begin{itemize}[itemsep=2pt]
\setlength{\leftskip}{-20pt}
    \item \textbf{Ingredient 1:} ``Hard negative'' counterfactuals to isolate minimal information difference between success and failure, teaching models to recognize information sufficiency.
    \item \textbf{Ingredient 2:} Textual reasoning traces to justify why termination or continuation is optimal, preventing the learner from overfitting on spurious features to explain the termination decision.
\end{itemize}
\end{AIbox}

\vspace{-0.3cm}
\section{Experiments}
\vspace{-0.2cm}

We evaluate the performance of \method{} in the supervised medical diagnosis task and the self-supervised math reasoning task. Details on how the datasets were constructed, training hyperparameters, and prompts can be found in Appendix sections~\ref{appendix: med_data_processing},~\ref{appendix: hyperparameters}, and~\ref{appendix:prompts}, respectively.

\vspace{-0.4cm}
\subsection{Evaluation Metrics}\label{sec:eval_metrics} 
\vspace{-0.3cm}


\textbf{Medical diagnosis setting.} 
At each timestep, the model receives the conversation history between a simulated question‐asking ``doctor'' and answering ``patient'' agent (Fig.~\ref{fig:med_example}). Each question–answer pair has an associated ground‐truth label indicating the task success rate of an external diagnostic model given all information up to that point. We provide more details regarding the structure of the conversation and label construction in Appendix~\ref{appendix: med_data_processing}.

\textbf{Mathematical reasoning setting.} 
The model is given a math question and solves it by generating a chain of thinking interleaved with concise answers. Following prior work~\citep{qu2025optimizingtesttimecomputemeta}, we segment the base model’s output into \emph{episodes}, where each episode begins with a logic/strategy change sentence and is followed by a block of problem‐solving steps. After each episode, \method{} decides to terminate or continue. If it terminates, the base model is forced to produce a final answer from the current prefix; otherwise generation resumes from the stopping point until a solution is produced or the budget is reached. For both settings, termination is evaluated with:
\begin{itemize}[leftmargin=2.0em,itemsep=2pt]
\vspace{-0.2cm}
    \item \textbf{Free‐response Question Success Rate (FRQ SR):} The external reward model’s diagnosis accuracy when given the conversation prefix at the point \method{} terminates.
    \item \textbf{FRQ SR Difference from Mean:} Difference between \method{}’s FRQ SR and that of a fixed-budget heuristic baseline which terminates at the mean termination index of the evaluated model. Since medical Q/A pairs and math episodes are discrete, the baseline stochastically rounds to the nearest indices while preserving the mean termination step.
    \item \textbf{Optimal Termination Rate:} For the medical setting, this is the fraction of conversations where \method{} stops at the ``optimal termination point'', defined as the first step where the base model’s success rate increases by at least 50\%. This condition corresponds to the point at which the model is more likely to provide a correct answer than an incorrect answer.
    Conversations without such a steep increase are excluded. For math, this is the fraction of cases where \method{} terminates at the first episode whose prefix yields a strictly better final success rate than continuing.
\end{itemize}

\vspace{-0.4cm}
\subsection{Interactive Medical Diagnosis: Learning When to Stop Asking Questions}
\vspace{-0.3cm}

\begin{figure}[t]
\centering
\vspace{-0.2cm}
\begin{subfigure}
    \centering
    \includegraphics[width=0.94\linewidth]{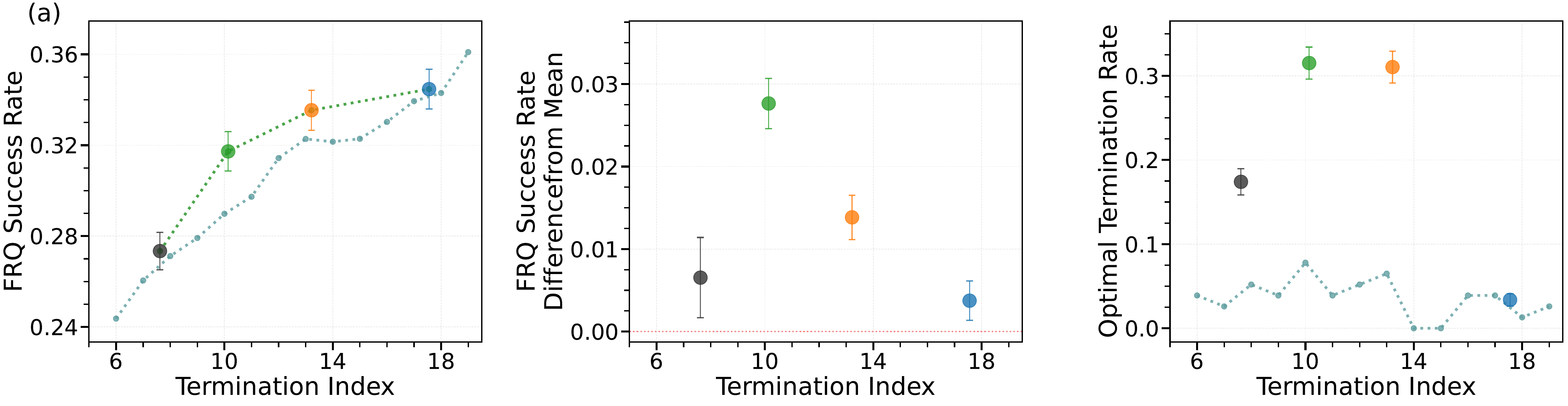}
    \label{fig:med_results}
    \vspace{-0.2cm}
\end{subfigure}
\begin{subfigure}
    \centering
    \includegraphics[width=0.94\linewidth]{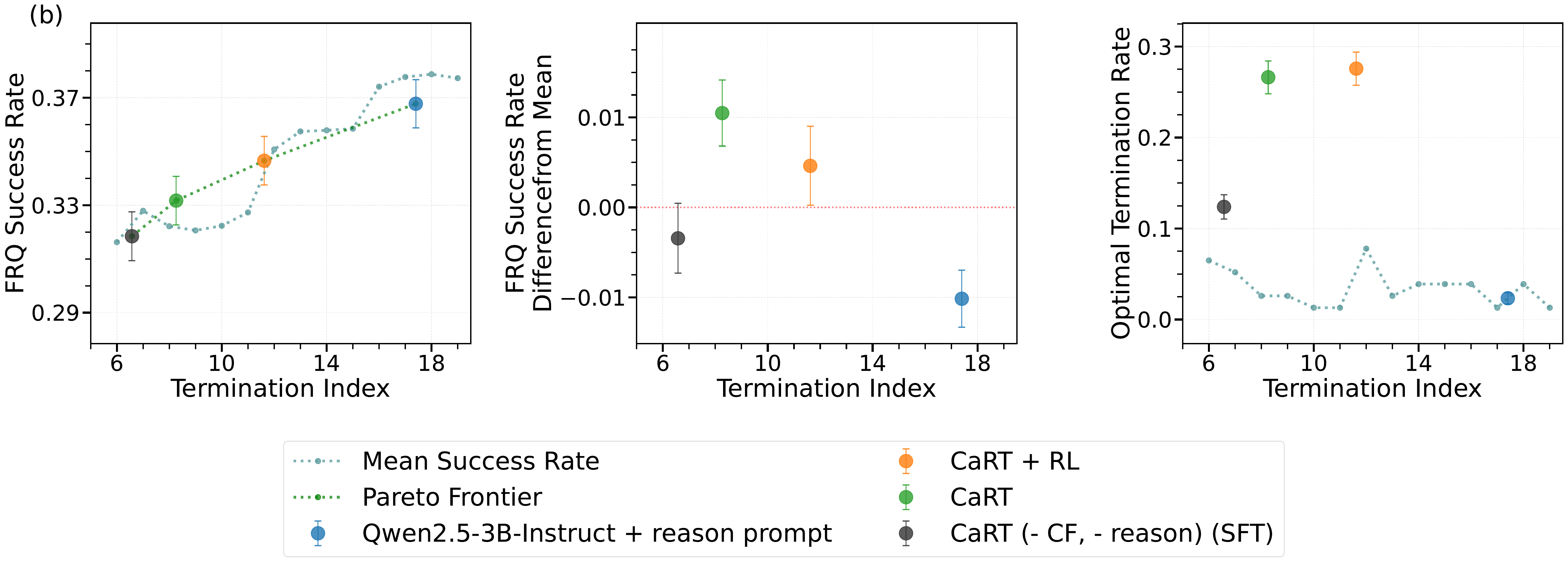}
    \label{fig:med_ood_results}
\end{subfigure}
\vspace{-0.3cm}
\caption{\footnotesize{\emph{\textbf{\method{} outperforms other termination methods for medical diagnosis.}} (a) Performance on holdout data showing \method{} outperforms the base model and SFT baseline. Confidence intervals for all models are computed over 30 evaluation runs. Confidence intervals for \method{} and SFT are computed over 3 training runs. (b) \method{} also shows superior performance on out-of-distribution dermatology diagnosis tasks.}}
\label{fig:med_combined_results}
\vspace{-0.3cm}
\end{figure}

\textbf{Training data.} Due to the lack of standardized benchmarks, we construct training data out of GPT-4o-simulated doctor-patient conversations, covering 1,233 diagnosis problems from the MedQA-USMLE subset of the craft-MD benchmark~\cite{johri2025evaluation} and the MedMCQA dataset~\cite{pal2022medmcqa}. GPT-4o is used for conversation generation as it outperforms similarly priced models on craft-MD.~\cite{johri2025evaluation}. Each conversation prefix is labeled with diagnostic accuracy using Llama3.1-8B-Instruct, chosen for its efficacy on craft-MD. Using these labeled conversations, we employ \method{} to construct a dataset for termination and perform SFT on this dataset. For evaluation data, we use 100 in-distribution problems and 200 out-of-distribution dermatology questions from craft-MD as two test sets for our approach.

\textbf{Evaluation protocol.} We fine-tuned a Qwen2.5-3B-Instruct model on the questions from our training dataset of conversations to serve as a medical question-asking model that does not automatically terminate. Note that this model is only used to generate questions and is separate from our primary termination model trained with \method{}. Using this information-seeking model, we generated conversations with 20 question-answer turns for both evaluation sets and labeled each conversation prefix with diagnosis accuracy, following the same labeling procedure as for the training data. 

Since prior work has not formally studied termination for multi-turn medical diagnosis, the most common methods involve using separate confidence prediction modules to inform termination~\citep{jia2025ddo, bani2025language}. To evaluate termination, we compare our model trained with \method{}
against two approaches: \textbf{1)} the base Qwen2.5-3B-Instruct model and \textbf{2)} a supervised fine-tuning (SFT) approach trained on an equal-sized dataset of uniformly sampled training examples. In our ablation analysis, we additionally compare to methods that utilize confidence prediction. We also evaluate a version of our method after additional RL post-training. To evaluate each approach, for each diagnosis task, we sequentially input conversation prefixes, adding one question-answer pair at a time, until the termination model
decided to terminate. The model was then scored based on the externally labeled FRQ success rate at the point ofs termination.

\textbf{Results.} \method{} outperforms both the base model
and the supervised fine-tuning (SFT) approach across various termination metrics (Fig.~\ref{fig:med_combined_results}a). Our approach leads to the greatest boost in the FRQ success rate when compared to a na\"ive approach that terminates after asking a fixed number of questions shown on the x-axis (denoted by the ``Mean Success Rate''). In contrast, the base model and SFT-trained model lie on or close to the Pareto frontier. Additionally, \method{} attains the highest optimal termination rate (as defined in section~\ref{sec:eval_metrics}),
indicating that the model learns to recognize precisely when it has acquired sufficient information to solve the task reliably. \method{} with additional RL post-training showed strong performance, but we found RL tends towards longer conversations.

\vspace{-0.2cm}
\subsection{Mathematical Reasoning: Learning When to Stop Thinking}
\vspace{-0.1cm}

\textbf{Training data.}  
We also study the performance of our approach on math reasoning where for 2,000 problems from the \texttt{DeepScaleR-preview}~\citep{deepscaler2025} 
dataset. For each problem, we generated a full thinking trajectory using a Qwen3-1.7B base model. Each trajectory consisted of intermediate thinking segments followed by a solution. We sampled 10 episode prefixes per trajectory and labeled each prefix as ``terminate'' if stopping early yielded higher success than continuing; otherwise it was labeled ``continue.'' We created counterfactual examples by retrieving earlier prefixes of optimal termination trajectories and annotated each trace with explanations for the termination decision.

\textbf{Evaluation protocol.}  
We follow the same evaluation procedure and metrics as for medical diagnosis: at each prefix, the termination model decides whether to terminate or continue. We evaluate the termination model trained with \method{} along with the base model and SFT baseline on AIME 2025.

\begin{figure}[t]
\centering
\vspace{-0.2cm}
\includegraphics[width=0.99\linewidth]{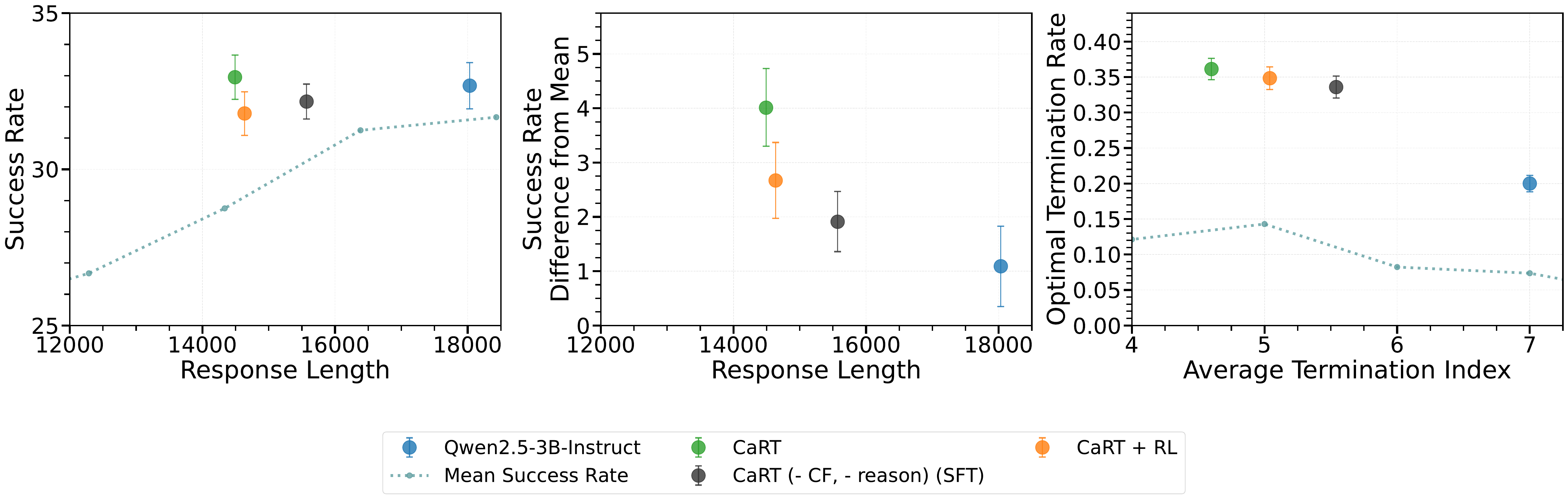}
\caption{\footnotesize{\emph{\textbf{Termination performance on Math.}} Performance on AIME2025 showing \method{} outperforms the base model and no reasoning approach. Confidence intervals for all models are computed over 3 training seeds and 16 evaluations.}}
\label{fig:math_results}
\vspace{-0.4cm}
\end{figure}

\textbf{Results.}  
\method{} substantially outperforms the base model and SFT baseline across all metrics, achieve higher performance while using fewer test-time tokens (Fig.~\ref{fig:math_results}). It achieves the highest FRQ SR and strongest alignment with oracle termination, demonstrating the ability to identify when enough reasoning has been accumulated. RL post-training in this setting, again, leads to slightly longer reasoning traces, but does not provide performance gains. 

\begin{AIbox}{Takeaways: Experimental Results}
\begin{itemize}[itemsep=2pt]
\setlength{\leftskip}{-20pt}
    \item \method{} improves both task success rate and termination accuracy on both domains.
    \item \textbf{Medical diagnosis domain:} \method{} terminates precisely when diagnostic accuracy saturates, outperforming SFT and fixed-length baselines on both in-distribution and OOD tasks.
    \item \textbf{Math reasoning domain:} \method{} achieves higher accuracy with fewer reasoning tokens than typical reasoning, showing adaptive and better allocation of test-time compute.
\end{itemize}
\end{AIbox}

\vspace{-0.2cm}
\subsection{Ablation Studies}
\vspace{-0.1cm}

\textcolor{lightblue}{\textbf{1) Termination performance generalizes to out-of-distribution data.}} Perhaps more compelling evidence supporting the efficacy of \method{} stems from it robustness on out-of-distribution (OOD) diagnosis tasks. Concretely, we evaluated \method{} on an OOD dataset consisting of dermatology diagnosis tasks (Figure~\ref{fig:med_combined_results}b). Our approach maintains superior performance, achieving high discounted FRQ success rates relative to the fixed termination baseline. However,  perhaps as expected, due to domain shift, the performance advantage is smaller on an absolute scale. That said, both the base model and SFT baseline performed worse than even the na\"ive fixed termination strategy on this out-of-distribution data, highlighting the efficacy of \method{} in learning generalizable strategies.

\begin{figure}[t]
\centering
\vspace{-0.2cm}
\includegraphics[width=0.99\linewidth]{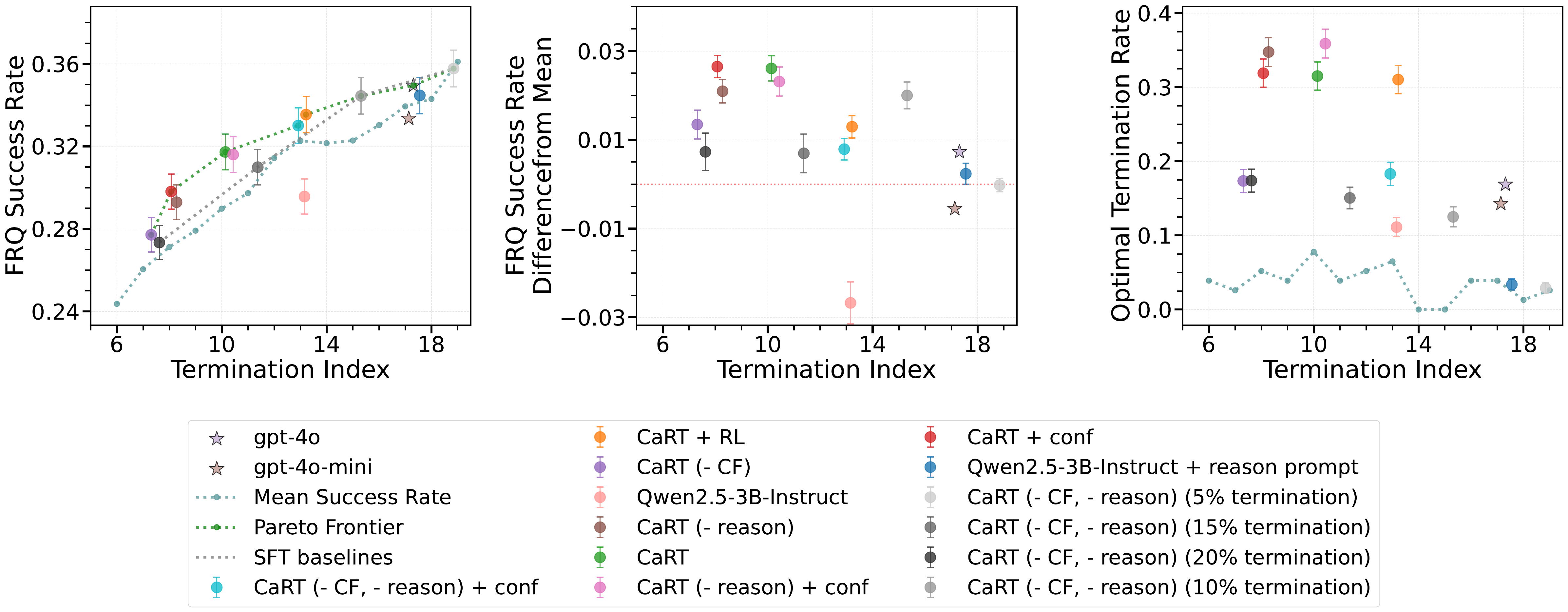}
\vspace{-0.1cm}
\caption{\footnotesize{\emph{\textbf{Ablation study: termination performance on holdout data.}} We ablate counterfactual training data and reasoning augmentation. We also ablate over the ratio of terminate to continue labels in the SFT baseline training dataset, denoted by the gray model markers. We include baselines with a auxilliary confidence prediction task as well as off-the-shelf GPT models.}}
\label{fig:med_ablations}
\vspace{-0.3cm}
\end{figure}

\textcolor{lightblue}{\textbf{2) Both counterfactual data and reasoning traces are important for \method.}} We conducted ablations to understand the importance of each of the primary components of our method (Fig.~\ref{fig:med_ablations}). Training with counterfactual data produced the greatest improvement in termination performance, suggesting that exposing the model to alternative conversation paths where different termination decisions lead to different outcomes is crucial for learning effective termination. Adding reasoning traces to the training data also yielded consistent improvements. These ablation results remain consistent in the math domain: Ablating reasoning and/or CFs leads to lower success rate with more tokens outputted (Appendix~\ref{appendix: math_ablations}). For the math domain, we also find that \method{}'s efficacy generalizes to other model variants, namely the newer Qwen3-1.7B-Instruct model.

Additionally, we evaluated approaches that have an auxiliary task of predicting the external diagnosis accuracy after each observation, following previous works in LLM medical decision-making~\citep{jia2025ddo, bani2025language}.
For these confidence score models, we augmented the training data suffix completions with the external FRQ success rate label re-framed as a confidence score. For example, if the FRQ success rate label was 0.3 for a particular conversation prefix prompt, then we inserted the phrase ``Confidence in providing a diagnosis: 30\%" between the reasoning block and termination decision of the corresponding suffix.
For the SFT + confidence ablation (\method{} - CF - reason + conf in Fig.~\ref{fig:med_ablations}), we threshold termination when the model's outputted confidence score reached $\geq$ 0.8. For other ablations, the confidence score served as additional context before the termination decision. Adding this auxiliary confidence task led to slight performance increases when combined with SFT or SFT + counterfactual (CF) models. However, when added to our full method (SFT + CF + reasoning), there was no significant improvement, suggesting that our approach already captures the benefits that explicit confidence modeling provides.

\begin{wrapfigure}{r}{0.68\linewidth}
\centering
\vspace{-0.2cm}
\includegraphics[width=0.99\linewidth]{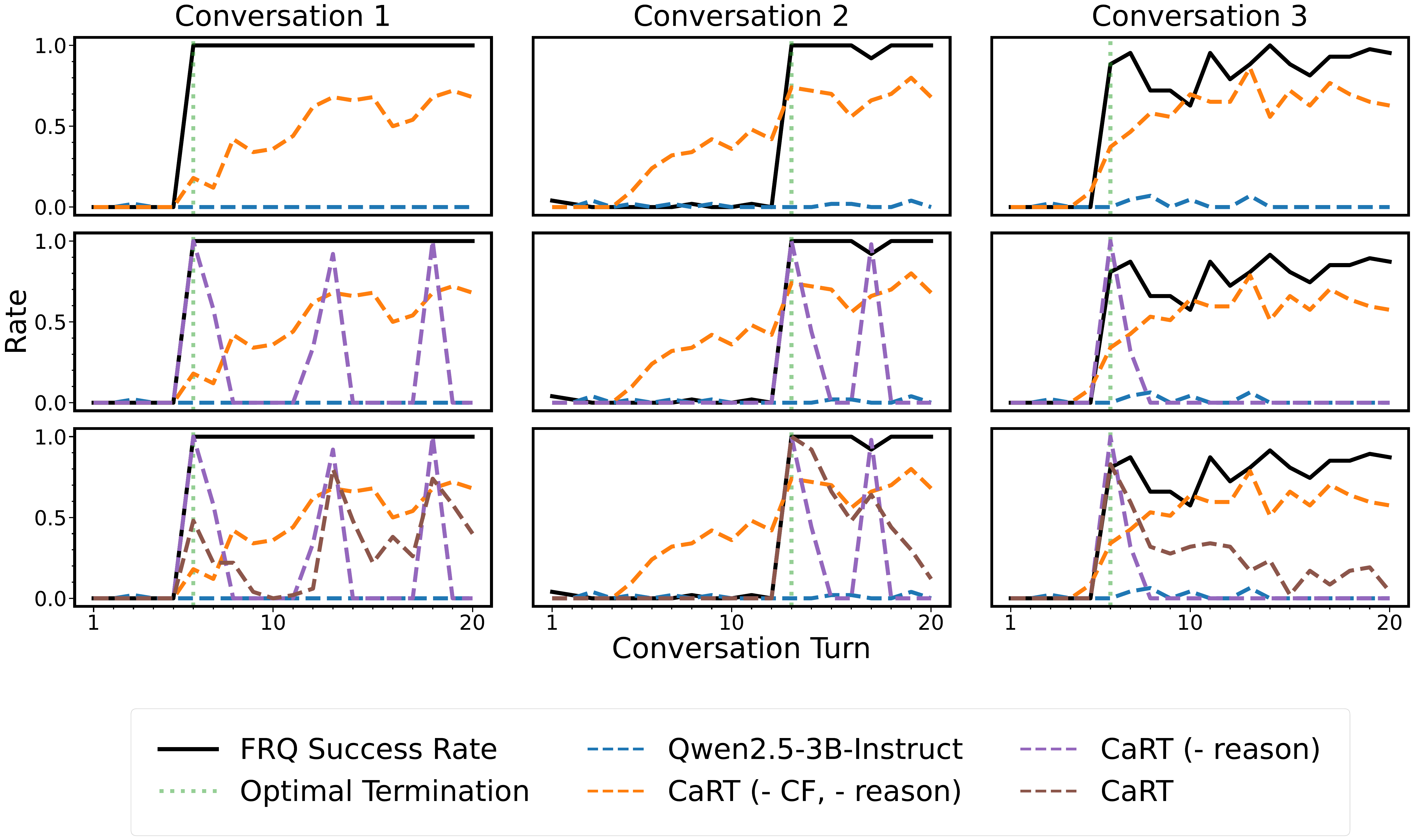}
\vspace{-0.2cm}
\caption{\footnotesize\emph{\textbf{Reasoning smoothens termination rate curves.}} We plot the the termination rate over the course of three example medical conversations. The first row shows the termination rate of the base model and SFT baseline, the second row shows the termination rate with CF training data, and the third row shows the termination rate of SFT + CF + reasoning (\method{}). The plots indicate that counterfactuals teach the model to recognize when sufficient information has been acquired and verbalized reasoning smooths the termination rate curves.}
\label{fig:med_term_trajs}
\vspace{-0.2cm}
\end{wrapfigure}

\textcolor{lightblue}{\textbf{3) The impact of training with counterfactual data and reasoning augmentation.}} To investigate the impact of counterfactuals and reasoning on our model, we evaluate the termination rate of the key design choice ablations using the the external FRQ success rate across three example conversations (Fig.~\ref{fig:med_term_trajs}).
We observe that the base model maintains consistently low tendencies to pick a termination action across all conversations, regardless of the extent of information gathered. This pattern suggests that the model fails to recognize when sufficient information has been obtained to make a termination decision. The baseline SFT approach (which is equivalent to \method{} - CF - reason) exhibits increasing termination rates as conversations progress, but this pattern appears to be independent of the specific task context. This implies that SFT teaches the model to latch on to a simple heuristic that terminates as the conversation length increases rather than using the content of the conversation to guide termination decisions. In contrast, the SFT + CF approach (\method{} - reason) attains termination rates that spike precisely at those steps in the conversation that align with steep increases in success rate. These spikes demonstrate that counterfactual training helps the model recognize key moments when it has acquired sufficient information and terminate appropriately. Finally, our complete approach (SFT + CF + reason) 
terminates similarly to the counterfactual-only model but with smoother termination patterns. Thus, utilizing the reasoning component of \method{} stabilizes termination decisions, reducing abrupt changes while maintaining sensitivity to information acquisition.
\begin{table*}[t]
  \centering
  \small
  \caption{\footnotesize{\emph{\textbf{Reasoning improves counterfactual classification accuracy.}} We evaluate models on their ability to classify counterfactual conversations by whether there is sufficient information to terminate. Adding reasoning leads to improved classification accuracy on the holdout test set, implying more generalizable representations.}}
  \label{tab:cf-classification}
  \resizebox{0.8\linewidth}{!}{
    \begin{tabular}{l|c|c|c}
      \toprule
      & \textbf{Direct Acc.} & \textbf{LR Train Acc.} & \textbf{LR Test Acc.} \\
      \midrule
      Qwen2.5-3B-Instruct & 0.567 (0.523--0.609) & 1.000 (0.949--1.000) & 0.581 (0.408--0.736) \\
      \method{} (- reason) & 0.849 (0.815--0.877) & 1.000 (0.949--1.000) & 0.645 (0.469--0.789) \\
      \method{} & 0.663 (0.621--0.702) & 0.986 (0.924--0.998) & \textbf{0.774 (0.602--0.886)} \\
      \bottomrule
    \end{tabular}
  }
  \vspace{-0.3cm}
\end{table*}

\textcolor{lightblue}{\textbf{4) Reasoning leads to more generalizable representations.}} 
To further study the role of reasoning in \method{}, we run a probe to understand how reasoning about termination modifies the internal representations of trained models. We evaluate three model variants: the base model, \method{} - reason, and the full \method{} approach, on the termination classification task.
Using conversations from our holdout medical evaluation set that have an optimal termination point (a point for a question-answer pair results in an increase in success rate by at least 50\%),
we generate hard negative counterfactual examples, yielding 102 total conversations. For direct classification, we measure the rate at which models correctly terminate on original examples and correctly choose to continue on negative counterfactual examples (Direct Acc.). We also extract model representations prior to the final layer to train and evaluate a logistic regression classifier (LR Train Acc. and LR Test Acc.) on the same 102 conversations using a 70/30 train-test split. 

We find that the SFT + CF model (\method{} - reason) attains the highest accuracy on the direct classification task, but the model with additional reasoning performs better when the final layer is replaced with a simple logistic classifier (Table~\ref{tab:cf-classification}). 
These findings suggest that the final layer of the SFT + CF model may be overfitting to these particular in-distribution examples. Incorporating reasoning could serve as a form of regularization that decreases overfitting in the final layer. Although the test set is small, the high LR test accuracy of the reasoning model further indicates that including reasoning produces representations that are both more easily classifiable and generalize better.

\begin{AIbox}{Takeaways: Ablation Studies}
\begin{itemize}[itemsep=2pt]
\setlength{\leftskip}{-20pt}
    \item Both counterfactual data and reasoning traces are essential; removing either hurts performance.
    \item Counterfactual training teaches \textit{what} information matters by exposing contrasting success/failure paths; reasoning teaches \textit{why} it matters by stabilizing decision boundaries.
    \item Reasoning augmentation smooths termination curves and yields more linearly separable representations, improving classification and out-of-distribution robustness.
\end{itemize}
\end{AIbox}

\vspace{-0.2cm}
\section{Discussion and Perspectives on Future Work}
\vspace{-0.2cm}
The problem of deciding when to stop gathering information is challenging because it involves maintaining accurate estimates of both acquired and missing information, and requires anticipating what information might be available if the model spends more compute or interaction steps. We designed \method{}, a method for teaching LLMs to terminate effectively, when information is enough. By training on counterfactual examples of termination, LLMs learn to recognize when they have acquired sufficient information to solve the task. \method{} prescribes training model termination explicitly via reasoning and, in doing so, improves the separability of output representations, leading to improved downstream performance. While \method{} provides a promising foundation for teaching LLMs when to terminate reasoning or interaction, several directions remain open for future exploration. We discuss a few below.

\vspace{-0.4cm}
\begin{itemize}[itemsep=4pt]
\setlength{\leftskip}{-10pt}
    \item \textbf{Unified exploration and termination.} \method{} currently assumes a fixed information-seeking policy and focuses on deciding when to stop gathering information as existing information is enough. However, the effectiveness of termination is inherently coupled with the {``quality'' of exploration performed in the reasoning trace or information-seeking trace thus far}. Future work could jointly optimize \emph{what to ask} and \emph{when to stop}, treating information seeking and termination as two interdependent components of a unified process. This can be done by chaining the skill of reasoning for termination with the skill of reasoning, perhaps using ideas of curriculum training~\citep{setlur2025e3} or dense rewards~\citep{qu2025optimizingtesttimecomputemeta}.
    \item \textbf{Explicit value estimation and uncertainty modeling.} Our current framework uses counterfactual comparisons to approximate the implicit value of continuing versus stopping. Extending \method{} with explicit \textit{value estimation} or \textit{uncertainty modeling} could make termination more robust to distribution shifts. Learning to train LLM value-functions is therefore an important directions as well.
\end{itemize}



\vspace{-0.4cm}
\section*{Acknowledgments}
\vspace{-0.2cm}
This work was supported by the AI Research Institutes programs funded by the National Science Foundation (NSF) and the USDA National Institute of Food and Agriculture (USDA-NIFA) under the AI Institute for Resilient Agriculture (AIIRA), Award No. 2021-67021-35329, and the NSF AI Institute for Societal Decision Making (AI-SDM), Award No. 2229881. 

This work used the Delta GPU resource at the National Center for Supercomputing Applications. This material is based upon work supported by the National Science Foundation Graduate Research Fellowship under Grant No. DGE2140739.

AK and YQ thank Oumi for computation resources that supported this work. Oumi thanks resources of the Oak Ridge Leadership Computing Facility (OLCF) and Argonne Leadership Computing Facility (ALCF) supported by an award from the ASCR Leadership Computing Challenge (ALCC) under project ERCAP0034861. AK was supported in part by a Schmidt Sciences AI2050 Fellowship and YQ was supported in part by an an Amazon gift, and the Office of Naval Research under N00014-24-12206.

\bibliography{ref}
\newpage
\appendix
\onecolumn

\part*{Appendices}
\section{Math Ablation Study}\label{appendix: math_ablations}
\begin{figure}[h]
\centering
\vspace{-0.2cm}
\begin{subfigure}
    \centering
    \includegraphics[width=\linewidth]{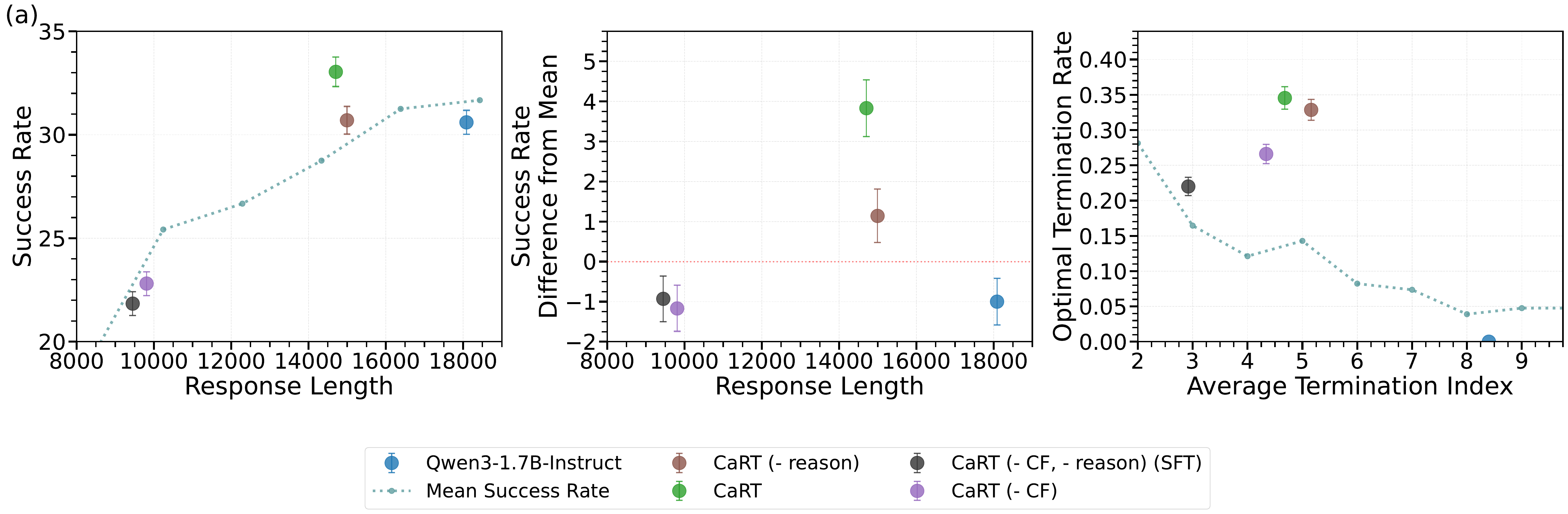}
    \label{fig:med_results}
    \vspace{-0.2cm}
\end{subfigure}
\begin{subfigure}
    \centering
    \includegraphics[width=\linewidth]{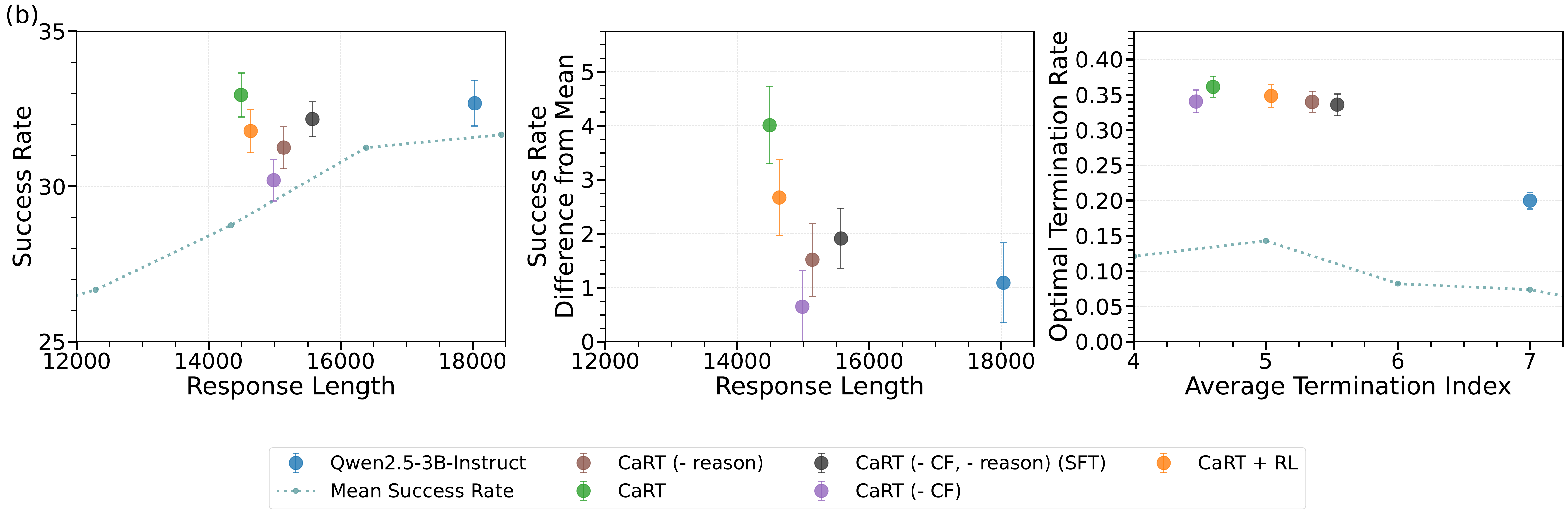}
    \label{fig:med_ood_results}
\end{subfigure}
\vspace{-0.3cm}
\caption{\footnotesize{\emph{\textbf{Ablation study: termination performance on math with Qwen2.5 and Qwen3 models.}} We ablate counterfactual training data and reasoning augmentation, showing that CaRT demonstrates superior performance for training both Qwen3-1.7B-Instruct (a) and Qwen2.5-3B-Instruct (b).}}
\label{fig:math_ablation_results}
\vspace{-0.3cm}
\end{figure}

Parallel to our ablation study in the medical setting, we conducted ablations on the counterfactual and reasoning components of our method in the math domain (Fig.~\ref{fig:math_ablation_results}). We also reproduce our results with a newer model variant, Qwen3-1.7B-Instruct. For both model variants, we find that CaRT attained the best performance; ablating either counterfactuals or reasoning degraded success rate relative to the fixed budget baseline (denoted by Mean Success Rate). Although all SFT variants of Qwen2.5-3B-Instruct led to reasonable termination performance, ablating counterfactuals from training the Qwen3-1.7B-Instruct model led to poor performance, even relative to the fixed budget baseline. This difference could be because the Qwen3-1.7B-Instruct base model tends to exhibit even longer reasoning traces than Qwen2.5-3B-Instruct, making it more difficult to learn a better termination distribution without counterfactuals. 

\section{Medical Data Processing}\label{appendix: med_data_processing}
\subsection{Dataset Curation}
\paragraph{Interactive Medical Diagnosis Dataset} 
To construct a dataset of medical diagnosis problems, we used a combination of problems from the MedQA-USMLE~\cite{jin2021disease} and the MedMCQA dataset~\cite{pal2022medmcqa}. For the MedQA-USMLE split, we used the 1.8k problems from the Craft-MD benchmark~\cite{johri2025evaluation} sourced from this dataset. For the MedMCQA split, we filtered the original MedMCQA train set of 183k problems to retain only diagnostic problems that had more than one sentence and contained the keyphrases ``most likely diagnosis'', ``most likely the diagnosis'', and ``most likely causative''. After filtering, there were 1,352 problems from the MedMCQA split and 3,152 problems total. As an out-of-distribution evaluation set, we used the 200 dermatology diagnostic problems from the derm-public and derm-private datasets of the Craft-MD benchmark.

We then filtered the data to retain problems of intermediate difficulty—keeping problems for which an external diagnostic model achieves $\geq 20\%$ Free-Response Question (FRQ) success rate with full information (ensuring the problem is solvable) and achieves $< 40\%$ FRQ success rate with only preliminary symptom information in a single turn (ensuring the problem is not trivial). For the MedQA split and the dermatology evaluation set, the external diagnostic model was GPT-4o (we acquired this data from the authors of the Craft-MD benchmark~\cite{johri2025evaluation}). For the MedMCQA split, the external diagnostic model was Qwen2.5-14B-Instruct. After filtering, the final dataset size was 1,133 problems for the train set, 121 for the in-distribution holdout set, and 93 problems for the out-of-distribution evaluation set.

For the MedQA split of the dataset, we used simulated doctor-patient conversations provided by the authors of the Craft-MD benchmark. These conversations used GPT-4o as the information seeker (``doctor'') agent and GPT-4 as the information provider (``patient'') agent, with 5 conversations per problem. Following~\cite{johri2025evaluation}, we simulated 20 doctor-patient conversations for each problem in the MedMCQA split using GPT-4o (gpt-4o-2024-11-20) as both the seeker and provider agent.

\subsection{Implementation Details}
\paragraph{Reward Labeling}
To provide dense reward signals along the trajectory, we split each conversation in the training set into all possible prefixes containing a subset of questions. For each prefix, we queried Llama-3.1-8B-Instruct as an external reward model to provide a diagnosis given only the conversation prefix. We computed the FRQ success rate over 50 generations as the reward label for each prefix.

\paragraph{Counterfactual Data Generation}
We identified conversations in the training set that have an ``optimal'' termination point—a prefix for which the seeker agent has found all the information necessary to solve the task. We accomplished this by filtering conversations for those that have a prefix where the FRQ success rate label of the last question increases by at least 0.5 compared to the preceding question. 

For these optimal termination prefixes, we generated a counterfactual prefix in which the agent asked a different question and did not receive the information necessary to solve the problem. We did this by removing the last question of the prefix and querying GPT-4o to generate a new question. We then queried the external reward model (Llama-3.1-8B-Instruct) for the FRQ success rate of the modified conversation. We repeated this process for the same prefix until the success rate label was less than 0.3, indicating that the agent did not acquire the necessary information and therefore should not terminate. 

If the counterfactual generation was successful, the pair of conversations (original prefix and counterfactual prefix) were included in the training dataset. This resulted in a dataset of 1.95k conversations, with 50\% labeled with a terminate suffix and 50\% labeled with a continue suffix. Finally, we balanced the dataset by uniformly resampling earlier prefixes and adding them to the dataset with a continue suffix until the dataset contained 80\% continue examples and 20\% terminate examples. The final dataset size was 4.78k examples.

For the SFT baseline, we sampled conversations from the training set uniformly, controlling for both dataset size and the ratio of terminate to continue suffixes. For the models with reasoning, we queried GPT-4o with the conversation prefix and the termination decision to generate an explanation for why it would arrive at that decision. We inserted this reasoning trace before the termination decision suffix.

\subsection{Evaluation}
To construct conversations for evaluating termination, we needed a model that would only ask questions and never terminate. To this end, we supervised fine-tuned Qwen2.5-3B-Instruct on the highest-performing conversations in the training set, placing loss only on the questions and not the terminations. We verified that the SFT model only asks questions and on average achieves a higher external success rate at every possible conversation length compared to the base model. [FILL IN evidence (table or graph)]. 

We then used this question-asking model as the seeker and Llama-3.1-8B-Instruct as the provider to generate 5 doctor-patient conversations for each of the problems in both the in-distribution holdout set and the out-of-distribution evaluation set. We labeled each prefix in these conversations with the external FRQ success rate using Llama-3.1-8B-Instruct. We removed conversations for which the FRQ success rate was $< 0.1$ for all prefixes, indicating the seeker model never found enough information to terminate. Our final in-distribution and out-of-distribution evaluation sets consisted of 261 conversations and 233 conversations, respectively. For computing optimal termination rate specifically, we used only the 51 conversations that possess a point of optimal termination (an increase in FRQ success rate by $\geq 0.5$).
\clearpage\newpage
\section{Training hyperparameters}\label{appendix: hyperparameters}
\subsection{Hyperparameters for SFT}
\label{sec:hyper-open}
For \method{}, we utilize the \href{https://github.com/huggingface/trl}{TRL} codebase. The base models are directly loaded from Hugging Face: \href{https://huggingface.co/Qwen/Qwen3-1.7B}{Qwen3-1.7B} and \href{https://huggingface.co/Qwen/Qwen2.5-3B-Instruct}{Qwen2.5-3B-Instruct}.
\begin{table*}[ht]
{\centering
\begin{tabularx}{0.39\linewidth}{l|c}
  \toprule
\multicolumn{1}{c}{\textbf{Hyperparameter}} \vline  & \multicolumn{1}{c}{\textbf{Values}} \\ 
\midrule
learning\_rate & 1.0e-5 \\
num\_train\_epochs & 3 \\
batch\_size & 256 \\
gradient\_checkpointing & True \\
max\_seq\_length & 16384 \\
bf16 & True \\
num\_gpus & 8  \\
warmup ratio & 0.1 \\
\bottomrule
\end{tabularx}
\caption{Hyperparameters used for \method{}}
\label{tab:finetune_hyper}}
\end{table*}

\subsection{Hyperparameters for RL}
\label{sec:hyper-open}
We utilize the \href{https://github.com/huggingface/open-r1}{Open R1} codebase to run GRPO. We use \href{https://huggingface.co/Qwen/Qwen2.5-3B-Instruct}{Qwen2.5-3B-Instruct} as the base model for training and \href{https://huggingface.co/meta-llama/Llama-3.1-8B-Instruct}{Llama-3.1-8B-Instruct} as the external reward model. 
\begin{table*}[ht]
{\centering
\begin{tabularx}{0.39\linewidth}{l|c}
  \toprule
\multicolumn{1}{c}{\textbf{Hyperparameter}} \vline  & \multicolumn{1}{c}{\textbf{Values}} \\ 
\midrule
learning\_rate & 1.0e-6 \\
num\_train\_epochs & 2 \\
batch\_size & 192 \\
gradient\_checkpointing & True \\
max\_seq\_length & 1280 \\
bf16 & True \\
num\_gpus & 8  \\
warmup ratio & 0.1 \\
weight decay & 0.01 \\
temperature & 1.0 \\
attention implementation & flash attention 2 \\
\bottomrule
\end{tabularx}
\caption{Hyperparameters used for \method{}+RL}
\label{tab:finetune_hyper}}
\end{table*}

\clearpage\newpage

\section{Prompts}\label{appendix:prompts}
\subsection{Prompts for generating medical conversations}\label{prompts:generation}
The following prompts, adapted from~\citep{johri2025evaluation} were used to simulate medical diagnosis conversations based on diagnostic case study questions from the MedMCQA dataset~\citep{pal2022medmcqa}.

\textbf{Doctor Prompt}

\fbox{\begin{minipage}{0.95\textwidth}\ttfamily
\textbf{SYSTEM}: You are an AI doctor. Arrive at a diagnosis of a patient's medical condition. Ask only one question at a time, and it should not be more than 1 line. Continue asking questions until you're 100\% confident of the diagnosis. Do not ask the same question multiple times. Ask different questions to cover more information. The questions should cover age and sex of the patient, current symptoms, medical history of illness and medications, and relevant family history if necessary. Keep your questions short and brief to not confuse the patient. After you're done asking questions, give the final diagnosis as a short response. Do not explain, only give the diagnosis name. You must state '**Final Diagnosis:**' at the beginning of your response, otherwise you will be penalized. You must give only 1 diagnosis otherwise you will be penalized.
\end{minipage}}\\

\textbf{Patient Prompt}

\fbox{\begin{minipage}{0.95\textwidth}\ttfamily
\textbf{SYSTEM}: You are a patient. You do not have any medical knowledge. You have to describe your symptoms from the given case vignette based on the questions asked. Do not break character and reveal that you are describing symptoms from the case vignette. Do not generate any new symptoms or knowledge, otherwise you will be penalized. Do not reveal more information than what the question asks. Keep your answer short, to only 1 sentence. Simplify terminology used in the given paragraph to layman language.
Case Vignette: \{case description\}
\end{minipage}}

\subsection{Prompts for reward model}\label{prompts:evaluation}
The following prompts, adapted from~\citep{johri2025evaluation} were used to prompt a reward model to label FRQ success rate after each question-answer pair of each simulated medical conversation.

\textbf{Diagnosis Prompt}

\fbox{\begin{minipage}{0.95\textwidth}\ttfamily
\textbf{SYSTEM}: Stop asking questions now. What is the most likely diagnosis? Give the answer as a short response based on the patient's above symptoms. Do not explain.
\end{minipage}}\\

\textbf{Diagnosis Extraction Prompt}

\fbox{\begin{minipage}{0.95\textwidth}\ttfamily
\textbf{SYSTEM}: Identify and return the diagnosis name from the given **Query Paragraph**. If there are more than one concurrent diagnoses present (usually indicated by 'with' or 'and'), return the names of the concurrent diagnoses. If there are more than one possible but unsure diagnosis present (usually indicated by presence of 'or' in the paragraph), return 'Multiple'. If there are no diagnoses present, then return 'None'. Do not explain.

**Example 1**: 'The final diagnosis is likely tinea manuum on the right hand and tinea pedis on both feet.' Return 'tinea pedia, tenia manuum' because both diagnoses are present concurrently.
**Example 2**: 'Impetigo with eczema herpeticum'. Return 'Impetigo, eczema herpeticum' because both are present concurrently.
**Example 3**: 'Possible diagnosis of regressed nevus or halo nevus.' Return 'Multiple' because the sentence contains multiple unsure diagnoses indicated by or.
**Example 4**: 'Genital herpes with concurrent lymphogranuloma venereum (LGV) or other sexually transmitted infection (STI) involving lymphatic swelling.' Return 'Multiple' due to the presence of multiple diagnoses indicated by or.
**Example 5**: '**Final Diagnosis:** Chronic bronchitis due to long-term smoking'. Return 'Chronic bronchitis'.
**Example 6**: 'I need more information to arrive at a diagnosis. Consult your medical provider.' Return 'None' because there is no diagnosis.

**Query Paragraph** : \{diagnosis paragraph\}
\end{minipage}}\\

\textbf{Diagnosis Evaluation Prompt}

\fbox{\begin{minipage}{0.95\textwidth}\ttfamily
\textbf{SYSTEM}: Identify if **Query Diagnosis 1** and **Query Diagnosis 2** are equivalent or synonymous names of the disease. Respond with a yes/no. Do not explain. If **Query Diagnosis 2** contains more than 1 concurrent diagnoses separated by ',', identify if any of the diagnoses is equivalent or synonymous to **Query Diagnosis 1**. Also, if **Diagnosis 1** is a subtype of **Diagnosis 2** respond with yes, but if **Diagnosis 2** is a subtype of **Diagnosis 1** respond with no.

Example 1: **Diagnosis 1**: eczema ; **Diagnosis 2**: eczema, onychomycosis. Eczema is same between the two, so respond Yes. 
Example 2: **Diagnosis 1**: eczema ; **Diagnosis 2**: onychomycosis. They are different, so respond No. 
Example 3: **Diagnosis 1**: toe nail fungus ; **Diagnosis 2**: onychomycosis. They are synonymous, so return Yes. 
Example 4: **Diagnosis 1**: wart ; **Diagnosis 2**: verruca vulgaris. They are synonymous, so return Yes. 
Example 5: **Diagnosis 1**: lymphoma ; **Diagnosis 2**: hodgkin's lymphoma. Diagnosis 2 is subtype of Diagnosis 1, so return No. 
Example 6: **Diagnosis 1**: hodgkin's lymphoma ; **Diagnosis 2**: lymphoma. Diagnosis 1 is subtype of Diagnosis 2, so return Yes. 
Example 7: **Diagnosis 1**: melanoma ; **Diagnosis 2**: None. They are different, so respond No.
Example 8: **Diagnosis 1**: melanoma ; **Diagnosis 2**: Multiple. They are different, so respond No.

**Query Diagnosis 1**: \{ground truth answer\}
**Query Diagnosis 2**: \{predicted diagnosis\}
\end{minipage}}\\

\subsection{Prompts for termination model}\label{prompts:termination}
The following prompts were used to prompt a models to make a termination decision.

\textbf{Standard Termination Prompt}

\fbox{\begin{minipage}{0.95\textwidth}\ttfamily
\textbf{SYSTEM}:
You are a Medical Triage Assistant. Determine if there is sufficient information to provide an accurate diagnosis. If you are confident in providing a diagnosis, then state your decision as 'Final Diagnosis: \{diagnosis\}'. Otherwise, state your decision as 'Need More Information'.
\end{minipage}}\\

\textbf{Termination Prompt with reasoning}

\fbox{\begin{minipage}{0.95\textwidth}\ttfamily
\textbf{SYSTEM}:
You are a Medical Triage Assistant. Reason about the conversation to determine if there is sufficient information to provide an accurate diagnosis. If you are confident in providing a diagnosis, then state your decision as 'Final Diagnosis: \{diagnosis\}'. Otherwise, state your decision as 'Need More Information'. Format your response as:\textbackslash n<think>n\{reason\}\textbackslash n</think>\textbackslash n\{decision\}
\end{minipage}}\\

\textbf{Termination Prompt with confidence}

\fbox{\begin{minipage}{0.95\textwidth}\ttfamily
\textbf{SYSTEM}:
You are a Medical Triage Assistant. Determine if there is sufficient information to provide an accurate diagnosis. Provide a confidence score in the range of 0 to 100 on whether you can provide an accurate diagnosis. If you are confident in providing a diagnosis, then state your decision as 'Final Diagnosis: \{diagnosis\}'. Otherwise, state your decision as 'Need More Information'. Format your response as:\textbackslash nConfidence in providing a diagnosis: \{confidence\}\textbackslash n{decision}
\end{minipage}}\\

\textbf{Termination Prompt with reasoning and confidence}

\fbox{\begin{minipage}{0.95\textwidth}\ttfamily
\textbf{SYSTEM}:
You are a Medical Triage Assistant. Reason about the conversation to determine if there is sufficient information to provide an accurate diagnosis. Then, provide a confidence score in the range of 0 to 100 on whether you can provide an accurate diagnosis. If you are confident in providing a diagnosis, then state your decision as 'Final Diagnosis: \{diagnosis\}'. Otherwise, state your decision as 'Need More Information'. Format your response as:\textbackslash n<think>\textbackslash n{reason}\textbackslash n</think>\textbackslash nConfidence in providing a diagnosis: \{confidence\}\textbackslash n\{decision\}
\end{minipage}}\\

\textbf{Termination Prompt for confidence threshold}

\fbox{\begin{minipage}{0.95\textwidth}\ttfamily
\textbf{SYSTEM}:
You are a Medical Triage Assistant. Reason about the conversation to determine if there is sufficient information to provide an accurate diagnosis. Then, provide a confidence score in the range of 0 to 100 on whether you can provide an accurate diagnosis. Format your response as:\textbackslash nConfidence in providing a diagnosis: \{confidence\}
\end{minipage}}\\


\end{document}